%% file: main.tex
\documentclass[10pt,twocolumn,letterpaper]{article}

\usepackage[pagenumbers]{wacv}

\input{preamble}

\definecolor{wacvblue}{rgb}{0.21,0.49,0.74}
\usepackage[pagebackref,breaklinks,colorlinks,allcolors=wacvblue]{hyperref}

\title{PRG-Fusion: Orchestrating Generative Priors with Reconstruction Evidence for Driving View Synthesis}

\author{
Sipeng He$^{1}$\thanks{This work was done during Sipeng He’s internship at Yootta.} \qquad Jialei Chen$^{2}$\thanks{Corresponding author. Email: \texttt{chenjialei@yootta.com}.} \qquad Zhen Fang$^{1}$ \qquad
Dongchun Ren$^{2}$ \qquad Feng Zhao$^{1}$\\
$^{1}$University of Science and Technology of China \qquad $^{2}$Yootta\\[0.5em]
}

\begin{document}
\maketitle
\input{sec/0_abstract}
\input{sec/1_intro}
\input{sec/2_related_work}
\input{sec/4_method}

\input{sec/5_experiments}
\input{sec/6_conclusion}

\clearpage

{
    \small
    \bibliographystyle{ieeenat_fullname}
    \bibliography{main}
}

\clearpage
\appendix
\input{sec/7_appendix}

\end{document}

%% file: preamble.tex
\newcounter{algorithm}
\newenvironment{paperalgorithm}[2]{%
  \refstepcounter{algorithm}%
  \label{#2}%
  \begin{figure}[t]
  \small
  \setlength{\parskip}{0pt}%
  \hrule height 0.7pt\vspace{3pt}
  \noindent\textbf{Algorithm \thealgorithm}\quad #1\par\vspace{3pt}
  \hrule height 0.35pt\vspace{3pt}
}{%
  \vspace{3pt}\hrule height 0.7pt
  \end{figure}
}
\newcommand{\algmeta}[2]{%
  \noindent\makebox[3.8em][l]{\textbf{#1:}}%
  \parbox[t]{\dimexpr\linewidth-3.8em\relax}{#2}\par
}
\newcommand{\algline}[2]{%
  \noindent\makebox[1.2em][r]{\scriptsize #1:}\hspace{0.8em}%
  \parbox[t]{\dimexpr\linewidth-2em\relax}{\strut#2}\par
}
\newcommand{\algindent}{\hspace*{1.15em}}
\newcommand{\algdoubleindent}{\hspace*{2.3em}}
\newcommand{\algkw}[1]{\textbf{#1}}

%% file: sec/0_abstract.tex
\begin{abstract}

Synthesizing photorealistic driving videos along specified trajectories is essential for scalable closed-loop simulation.
However, reconstructed scenes exhibit spatially varying degradation under viewpoint shifts, while generative refinement can alter reliable content and introduce geometric inconsistencies. 
We propose \mbox{\textbf{PRG-Fusion}}, a framework for driving view synthesis that uses reconstruction evidence to orchestrate generative priors across regions.
Our key insight is that reconstruction reliability and available support should jointly determine how previously unseen regions are generated and reliable content is preserved.
We derive pixel-wise Preserve, Repair, and Generate (PRG) labels from reconstruction evidence to distinguish reliable appearance, degraded but supported content, and regions with insufficient support.
At inference, these labels serve as a unified routing policy for region-aware spatiotemporal synthesis, orchestrating 3DGS appearance preservation, LiDAR-guided structural correction, and video-prior-driven content completion across Preserve, Repair, and Generate regions, respectively.
We then follow a two-stage training paradigm, first establishing geometric control from sparse LiDAR projections and subsequently learning appearance control from dense 3DGS renderings.
Extensive experiments on Waymo demonstrate that \mbox{\textbf{PRG-Fusion}} achieves state-of-the-art overall performance in novel trajectory video synthesis, with superior visual quality and geometric fidelity while maintaining competitive view consistency under large trajectory shifts.
Our project page is available at \url{https://pre-cici.github.io/PRG-Fusion/}.

\end{abstract}

%% file: sec/1_intro.tex
\begin{figure*}[t]
    \centering
    \includegraphics[width=\textwidth]{figures/figure1_overview.pdf}
    \caption{\textbf{Motivating example of PRG-Fusion.} A reconstructed driving scene is rendered along a target trajectory. The rendering and projected LiDAR provide complementary appearance and geometric evidence for assigning PRG regions. The highlighted crops compare the baseline rendering with the PRG-Fusion result in regions requiring generation or repair.}
    \label{fig:overview}
\end{figure*}

\section{Introduction}
\label{sec:introduction}

Developing reliable autonomous driving systems requires extensive training and evaluation across diverse driving scenarios. However, collecting real-world data that covers a wide range of driving behaviors and environmental conditions is costly and difficult to scale. Closed-loop simulation built from recorded driving logs offers a scalable way to reuse real-world observations for interactive policy evaluation~\cite{zhou2026hugsim,yang2023unisim}. As driving policies explore trajectories beyond those captured during data collection, the simulator must synthesize realistic observations from previously unobserved viewpoints while maintaining the appearance and geometric structure of the recorded scene. Achieving high-quality view synthesis under large trajectory shifts is therefore a key requirement for constructing faithful and controllable driving environments.

Neural scene representations, including NeRF and 3D Gaussian Splatting (3DGS)~\cite{mildenhall2020nerf,kerbl2023gaussians}, have enabled high-quality driving scene reconstruction. In particular, dynamic 3DGS methods~\cite{yan2024streetgaussians,chen2024omnire,huang2024s3gaussian,hess2025splatad} provide explicit scene representations suitable for efficient rendering. However, their quality remains constrained by the coverage of the recorded observations. As the target viewpoint moves away from the training trajectory, reconstruction degradation varies across the image: some regions retain accurate appearance, others exhibit distorted geometry or rendering artifacts, and newly exposed regions lack sufficient information for faithful reconstruction. These different local states require different treatments when extending the scene to novel trajectories, as illustrated in~\cref{fig:overview}.

Recent video generative models~\cite{gao2023magicdrive,gao2024vista,nvidia2026cosmos,guo2025dist4d,li2026omninwm,wang2024freevs,yan2024streetcrafter} provide rich visual and temporal priors for synthesizing realistic driving videos and completing content beyond observed views. However, sparse or incomplete conditioning can leave scene-specific geometry insufficiently constrained, allowing generated content to deviate from the recorded environment in vehicle appearance, lane structure, and spatial relationships. To better combine generative capabilities with reconstruction fidelity, FaithFusion~\cite{wang2025faithfusion} uses pixel-wise expected information gain (EIG) to guide generative refinement and weight reconstruction losses. Nevertheless, its formulation does not explicitly distinguish degraded regions that retain sufficient observational and geometric support for correction from regions that lack such support and require generative completion. Regions with similar unreliability may therefore benefit from different conditioning strategies. This motivates us to consider both reconstruction reliability and available support when guiding novel-trajectory video synthesis.

We propose \textbf{PRG-Fusion}, an evidence-guided framework for novel-trajectory driving video synthesis. We derive pixel-wise Preserve, Repair, and Generate (PRG) labels from target-view reconstruction reliability and available observational and geometric support. The labels distinguish reliable content, degraded regions with sufficient support for correction, and regions that require generative completion. During synthesis, PRG guides separate appearance and geometry control branches: reliable regions retain their 3DGS appearance, supported regions receive LiDAR-guided correction, and unsupported regions rely more on the video prior.

Since 3DGS renderings are highly correlated with the target video in appearance, directly learning from the two conditions jointly can cause the model to overly rely on the visual condition and reproduce its rendering artifacts, thereby weakening the geometric guidance from LiDAR. To address this issue, we adopt a two-stage progressive training strategy: we first establish stable geometric control using LiDAR projections, and then freeze the geometry branch while introducing the 3DGS visual condition, encouraging the two control signals to be learned in a complementary manner. Overall, our main contributions are:

\begin{itemize}
    \item We introduce PRG-Fusion, a region-aware video generation framework that explicitly formulates the spatially varying reliability of off-trajectory 3DGS renderings through pixel-wise Preserve, Repair, and Generate (PRG) labels, enabling reconstruction evidence to directly guide novel-trajectory video synthesis.
    \item We design a dual-branch control architecture with a two-stage progressive training strategy that effectively integrates dense 3DGS appearance and sparse LiDAR geometry, promoting complementary appearance and geometric control.
    \item Extensive experiments on the Waymo demonstrate that PRG-Fusion outperforms existing methods in novel-trajectory video synthesis, achieving superior visual quality, geometric fidelity, and view consistency.
\end{itemize}

%% file: sec/2_related_work.tex
\section{Related Work}
\label{sec:related-work}

\subsection{Driving Scene Reconstruction}
Recent advances in NeRF~\cite{mildenhall2020nerf,barron2022mipnerf360,muller2022instantngp,barron2023zipnerf} and 3DGS~\cite{kerbl2023gaussians,yu2023mipsplatting,lu2023scaffoldgs} have greatly accelerated the development of 3D scene reconstruction. For driving scene, StreetSurf~\cite{guo2023streetsurf} and EmerNeRF~\cite{yang2023emernerf} extend neural fields to large-scale dynamic street scenes through decomposed scene representations. Benefiting from explicit Gaussian primitives, 3DGS naturally facilitates the decoupled modeling of static backgrounds and dynamic foreground objects. Street Gaussians~\cite{yan2024streetgaussians} and DrivingGaussian~\cite{zhou2024drivinggaussian} leverage scene annotations to model dynamic actors compositionally, while S$^3$Gaussian~\cite{huang2024s3gaussian} and OmniRe~\cite{chen2024omnire} further reduce annotation requirements and support more general dynamic reconstruction. Despite these methods produce faithful renderings near recorded viewpoints, their quality remains bounded by camera coverage. Large trajectory deviations expose weakly constrained or unseen regions, causing geometric distortion, rendering artifacts, missing content, and temporal inconsistency.

\subsection{Generative Models for Autonomous Driving}
Video foundation models provide strong spatial and temporal priors for driving scene generation~\cite{blattmann2023videoldm,blattmann2023svd,wan2025,wu2025hunyuanvideo}. MagicDrive~\cite{gao2023magicdrive}, VISTA~\cite{gao2024vista}, Cosmos~\cite{nvidia2026cosmos}, and OmniNWM~\cite{li2026omninwm} further introduce trajectories, BEV maps, 3D bounding boxes, or navigation signals to enable controllable driving scene generation. For scene-specific novel trajectory synthesis, FreeVS~\cite{wang2024freevs} and StreetCrafter~\cite{yan2024streetcrafter} leverage scene priors projected into the target view or LiDAR geometry to guide generation, while DiST-4D~\cite{guo2025dist4d} incorporates metric depth for spatial control. However, such geometric conditions are often sparse or limited by occlusion, making it difficult to fully control the structure and appearance of the target view. As a result, regions with insufficient geometric support still rely heavily on generative priors, which can lead to content or geometric structures that are inconsistent with the real scene.

\subsection{Scene Reconstruction with Generative Priors}

Leveraging generative priors to improve scene reconstruction under limited observations~\cite{liu2026reconx,wu2024reconfusion,liu2024_3dgsenhancer,yu2025viewcrafter} has recently emerged as a promising direction, where generative models are used to restore or enhance unreliable novel-view renderings. 
In this line of work, ReconDreamer~\cite{ni2024recondreamer} and ReconDreamer++~\cite{zhao2025recondreamerplus} progressively expand reconstruction with world model priors; DriveX~\cite{yang2024drivex} iteratively co-optimizes video restoration and 3DGS;  Difix3D+~\cite{wu2025difix3d} distills diffusion-restored novel views into 3D representations; while FaithFusion~\cite{wang2025faithfusion} uses pixel-wise expected information gain to selectively guide generation and reconstruction updates.
Overall, these methods remain reconstruction-centric, treating generative priors mainly as a means to refine unreliable 3D content.
In contrast, PRG-Fusion directly converts reconstruction evidence into region-aware control signals for video generation, adaptively coordinating 3DGS appearance, LiDAR geometry, and generative priors according to the local scene state.

%% file: sec/4_method.tex
\begin{figure*}[t]
    \centering
    \includegraphics[width=\textwidth]{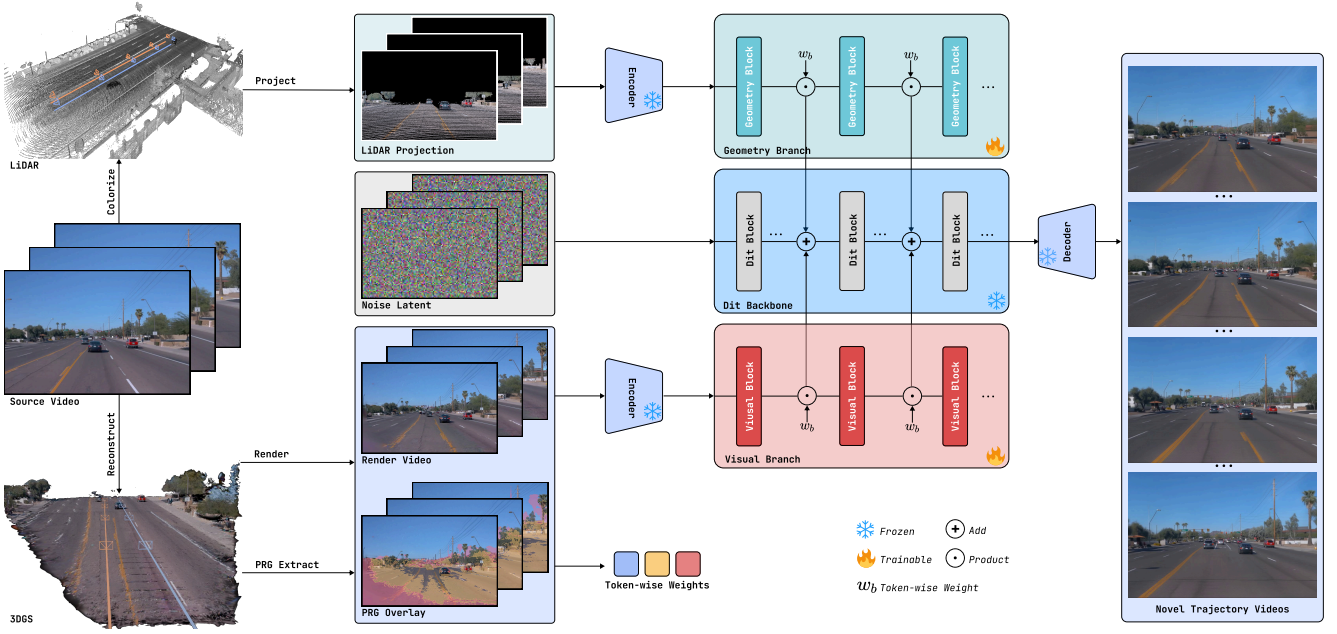}
    \caption{\textbf{PRG-Fusion pipeline.}
    Given a recorded driving log, we reconstruct a 3DGS scene and colorize the aggregated LiDAR sweeps. Along a target trajectory, 3DGS renderings and LiDAR projections form aligned visual and geometric conditions, while reconstruction reliability and support are used to derive pixel-wise PRG labels. The two conditions are encoded by separate control branches, and the PRG labels route their residuals into the frozen video DiT for novel-trajectory synthesis.}
    \label{fig:method-overview}
\end{figure*}

\section{Method}
\label{sec:method}
Given a recorded driving sequence and a target trajectory, PRG-Fusion synthesizes a temporally consistent video using 3DGS renderings, LiDAR projections, and a video generative prior. Pixel-wise Preserve, Repair, and Generate (PRG) labels derived from reconstruction evidence guide how the appearance and geometry control branches contribute across regions. Fig.~\ref{fig:method-overview} shows the overall framework.

\subsection{Preliminaries}
\label{sec:preliminaries}

\noindent\textbf{3D Gaussian Splatting.}
3DGS~\cite{kerbl2023gaussians} represents a scene with anisotropic Gaussians $\mathcal G=\{G_i\}_{i=1}^{N}$.
Each Gaussian has position $\mu_i\in\mathbb R^3$, rotation quaternion $q_i\in\mathbb R^4$, scale $\sigma_i\in\mathbb R^3$, opacity $o_i\in\mathbb R$, and spherical harmonic coefficients $\gamma_i$; we collect them as $\omega_i=(\mu_i,q_i,\sigma_i,o_i,\gamma_i)$.
After projection and alpha blending, the rendered pixel color is
\begin{equation}
    C=\sum_{i\in\mathcal M}
    c_i\alpha_i'
    \prod_{j=1}^{i-1}\left(1-\alpha_j'\right),
    \label{eq:3dgs-rendering}
\end{equation}
where $\mathcal M$ contains the Gaussians intersecting the pixel ray in depth order, and $\alpha_i'$ is determined by $o_i$ and the projected 2D Gaussian density.

\noindent\textbf{Conditional Generation with Flow Matching.}
Given a video $V$, a pretrained 3D VAE encoder $\mathcal{E}$ maps it into the latent space as $x_1=\mathcal{E}(V)$. Flow Matching~\cite{lipman2023flowmatching} learns a continuous transport from Gaussian noise $x_0\sim\mathcal{N}(0,I)$ to the data latent $x_1$ along the linear interpolation path
\begin{equation}
    x_t=(1-t)x_0+t x_1,
    \qquad t\in[0,1],
    \label{eq:flow-path}
\end{equation}
with the corresponding target velocity
\begin{equation}
    v_t=\frac{\mathrm d x_t}{\mathrm d t}
    =x_1-x_0.
    \label{eq:flow-velocity}
\end{equation}
Conditioned on $c$ and a uniformly sampled timestep $t\sim\mathcal{U}(0,1)$, the video model is trained to predict this velocity by minimizing
\begin{equation}
    \mathcal L_{\mathrm{FM}}
    =
    \mathbb E_{x_0,x_1,c,t}
    \left[
    \left\|
    v_\theta(x_t;c,t)-v_t
    \right\|_2^2
    \right],
    \label{eq:flow-matching}
\end{equation}
where $v_\theta$ denotes the conditional velocity predictor, and $c$ represents auxiliary conditions such as reference frames or other control signals.

\subsection{Reconstruction Evidence Extraction}
\label{sec:features}

Given a reconstructed 3DGS scene $\mathcal G=\{G_i\}_{i=1}^{N}$ and training views $\mathcal V^{\mathrm{train}}=\{V_k^{\mathrm{train}}\}_{k=1}^{K}$, and a target trajectory $\mathcal V^{\mathrm{tar}}=\{V_f^{\mathrm{tar}}\}_{f=1}^{T}$.
Our goal is to convert spatially varying reconstruction evidence into a pixel-wise PRG label map $L_f\in\{\mathrm P,\mathrm R,\mathrm G\}^{H\times W}$, where each target pixel is assigned to Preserve, Repair, or Generate.

\noindent\textbf{Trajectory unreliability.}
Following FaithFusion~\cite{wang2025faithfusion}, we use Expected Information Gain (EIG) to quantify the view-dependent unreliability of each Gaussian along the target trajectory.
The key idea is to compare parameter sensitivity between the training and target views.
For Gaussian parameters $\omega_i$ and differentiable renderer $\mathcal R$, we first accumulate the rendering-gradient responses over all training views:
\begin{equation}
    H_i^{\mathrm{train}}
    =\sum_{k=1}^{K}\left\lVert\nabla_{\omega_i}
    \mathcal R(V_k^{\mathrm{train}})\right\rVert_1.
    \label{eq:training-response}
\end{equation}
Each gradient norm measures how strongly one training rendering depends on $G_i$, and their sum represents the total training evidence constraining that Gaussian.
For target frame $f$, we evaluate the same parameter sensitivity without accumulation as $H_{i,f}^{\mathrm{tar}}=\left\lVert\nabla_{\omega_i}\mathcal R(V_f^{\mathrm{tar}})\right\rVert_1$.
The target response measures how strongly the novel view depends on $G_i$.
We normalize it by the accumulated training response to obtain the raw response ratio:
\begin{equation}
    e_{i,f}=\frac{H_{i,f}^{\mathrm{tar}}}{H_i^{\mathrm{train}}+\epsilon}.
    \label{eq:trajectory-unreliability}
\end{equation}
Here, $\epsilon$ stabilizes the denominator for weakly constrained Gaussians. We percentile-normalize $e_{i,f}$ within each target frame and aggregate the normalized responses over the full trajectory and local temporal windows to obtain $\bar u_i(f)$, as detailed in the supplementary material. Replacing the color attribute $c_i$ in Eq.~\eqref{eq:3dgs-rendering} with $\bar u_i(f)$ renders the pixel-wise unreliability map
\begin{equation}
    U_f(p)=\sum_{i\in\mathcal M}\bar u_i(f)\alpha_i'
    \prod_{j=1}^{i-1}\left(1-\alpha_j'\right).
    \label{eq:unreliability-projection}
\end{equation}

\noindent\textbf{Observation and geometric support.}
For each Gaussian $G_i$, we estimate its observation support based on visibility frequency and viewing-direction coverage. We first obtain a robust visibility score $v_i$ by logarithmically compressing and percentile-normalizing its visible-view count, while using the directional coverage $d_i$ as a modulation factor:
\begin{equation}
    s_i = v_i\big[(1-\lambda_d)+\lambda_d d_i\big].
    \label{eq:observation-support}
\end{equation}
Here, $s_i\in[0,1]$ denotes the observation support of $G_i$, and $\lambda_d$ controls the contribution of directional coverage.

However, observation statistics alone cannot determine whether the target-view geometry is supported by the recorded observations. We therefore measure depth consistency across valid training views:
\begin{equation}
    G_f(p)=
    \max_k
    \exp\!\left(
    -\frac{|D_{f\to k}(p)-D_k(p)|}
    {\tau_d\,\max(D_{f\to k}(p),D_k(p))}
    \right),
    \label{eq:geometric-support}
\end{equation}
where $D_{f\to k}$ denotes the projected target-view depth, $D_k$ the training-view depth, and $\tau_d$ controls the relative-depth tolerance.

Finally, we combine the projected observation support with the geometric support using the standard alpha-compositing weights:
\begin{equation}
    S_f(p)=G_f(p)
    \sum_{i\in\mathcal M}s_i\alpha_i'
    \prod_{j=1}^{i-1}\left(1-\alpha_j'\right).
    \label{eq:support-projection}
\end{equation}

\begin{figure}[t]
    \centering
    \includegraphics[width=\columnwidth]{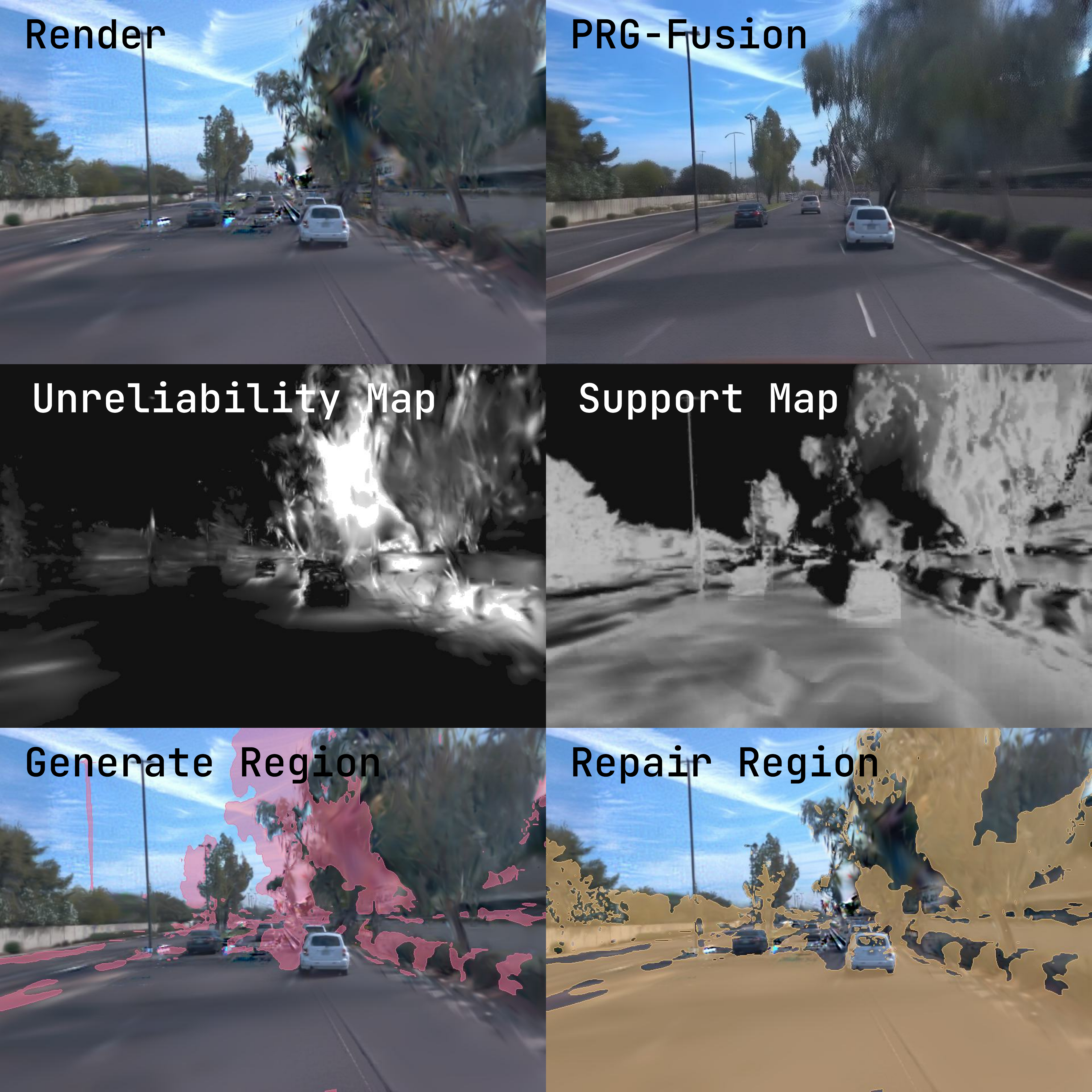}
    \caption{\textbf{Visualization of PRG evidence maps.} Unreliability and reconstruction support jointly determine Preserve, Repair, and Generate regions.}
    \label{fig:prg-visualization}
    \vspace{-8pt}
\end{figure}

\noindent\textbf{PRG label assignment.} 
As detailed in Alg.~\ref{alg:prg-extraction}, PRG assignment follows a hierarchical decision process. The unreliability map first identifies trustworthy pixels as Preserve. Among the remaining unreliable pixels, the support map further distinguishes well-supported regions as Repair and weakly supported regions as Generate. 
This assignment retains reliable reconstruction, enables evidence-guided correction where sufficient support exists, and delegates unsupported content to the video generative prior. 
The unreliability and support maps, together with the resulting Repair and Generate regions, are visualized in~\cref{fig:prg-visualization}.

\begin{paperalgorithm}{Reconstruction Evidence Extraction}{alg:prg-extraction}
\algmeta{Input}{3DGS $\mathcal G$; training views $\mathcal V^{\mathrm{train}}$; target trajectory $\mathcal V^{\mathrm{tar}}$; threshold $\tau_u,\tau_s$}
\algmeta{Output}{PRG maps $L=\{L_f\mid V_f^{\mathrm{tar}}\in\mathcal V^{\mathrm{tar}}\}$}
\vspace{2pt}
\algline{1}{$s_i\leftarrow\operatorname{Acc}(G_i,\mathcal V^{\mathrm{train}}),\quad \forall G_i\in\mathcal G$}
\algline{2}{$e_{i,f}\leftarrow\operatorname{EIG}(G_i,V_f^{\mathrm{tar}};\mathcal V^{\mathrm{train}}),\quad \forall i,f$}
\algline{3}{$\bar u_i(f)\leftarrow\operatorname{Aggregate}(\operatorname{Normalize}(\{e_{i,f}\}_{f=1}^{T})),\quad \forall i,f$}
\algline{4}{\algkw{for each} target view $V_f^{\mathrm{tar}}\in\mathcal V^{\mathrm{tar}}$ \algkw{do}}
\algline{5}{\algindent $U_f\leftarrow\operatorname{Render}_f(\bar u_i(f))$}
\algline{6}{\algindent $S_f\leftarrow\operatorname{Geo}_f\odot\operatorname{Render}_f(s_i)$}
\algline{7}{\algindent\algkw{for each} pixel $p$ \algkw{do}}
\algline{8}{\algdoubleindent $L_f(p)\leftarrow\mathrm P$ \algkw{if} $U_f(p)<\tau_u$}
\algline{9}{\algdoubleindent $L_f(p)\leftarrow\mathrm R$ \algkw{else if} $S_f(p)\geq\tau_s$}
\algline{10}{\algdoubleindent $L_f(p)\leftarrow\mathrm G$ \algkw{otherwise}}
\algline{11}{\algindent\algkw{end for}}
\algline{12}{\algkw{end for}}
\end{paperalgorithm}

\subsection{Dual-Branch Conditional Video Generation}
\label{sec:architecture}
Off-trajectory 3DGS renderings retain dense, scene-specific visual context, but their reliability degrades as the target trajectory departs from the training views. LiDAR projections, by contrast, provide reliable metric geometry, but their sparsity prevents them from describing complete visual appearance. This complementarity motivates our dual-branch conditioning design~\cite{jiang2025vace,xing2023dynamicrafter}: the visual branch captures dense scene appearance from 3DGS renderings, while the geometry branch injects LiDAR-based structural constraints for novel-trajectory synthesis.

\begin{figure*}[t]
    \centering
    \includegraphics[width=\textwidth]{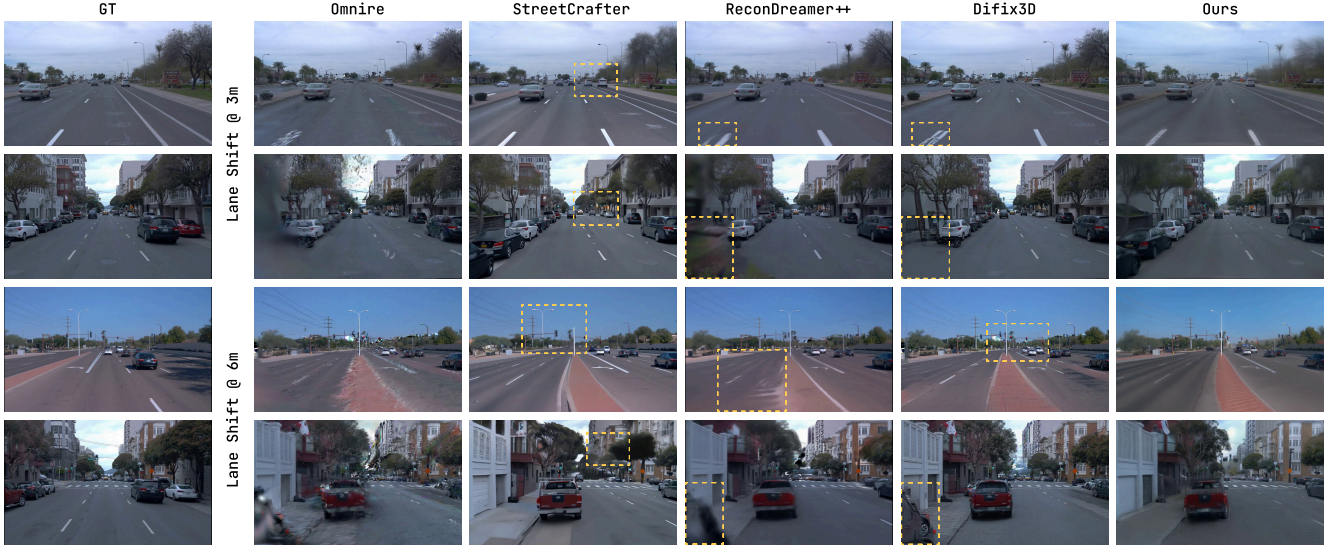}
    \caption{\textbf{Qualitative comparison under lateral trajectory shifts.} Note that GT is the reference view from the recorded trajectory, not the ground truth for the shifted target view.}
    \label{fig:qualitative-comparison}
    \vspace{-12pt}
\end{figure*}

\noindent\textbf{Data curation strategy.}
To train the dual-branch generative model, we construct strictly aligned triplets along the recorded trajectory, consisting of a degraded 3DGS rendering video $V_{\mathrm{render}}$, a colored LiDAR projection video $V_{\mathrm{lidar}}$, and the corresponding real video $V_{\mathrm{rgb}}$.
To construct $V_{\mathrm{render}}$, we sample structured degradation regions fixed in the world coordinate system and either remove image supervision entirely or retain only a small subset of frames for weak supervision, producing spatially coherent and cross-view-consistent under-constrained reconstructions.
For $V_{\mathrm{lidar}}$, following FreeVS~\cite{wang2024freevs}, we aggregate LiDAR points from $r$ neighboring frames and align dynamic points according to their object trajectories.
The aggregated point cloud is colored using synchronized camera images and projected into the target view, yielding a sparse three-channel RGB condition.

\noindent\textbf{Dual-branch architecture.}
We use a frozen video DiT as the generative backbone and construct a visual branch conditioned on 3DGS renderings and a geometry branch conditioned on LiDAR projections.
A shared frozen VAE encodes the aligned real, rendered, and LiDAR videos as $x_1=\mathcal E(V_{\mathrm{rgb}})$, $c_{\mathrm{render}}=\mathcal E(V_{\mathrm{render}})$, and $c_{\mathrm{lidar}}=\mathcal E(V_{\mathrm{lidar}})$.
The two condition latents are independently mapped to the video-token grid by their control embedders and passed to the corresponding ControlDiT blocks.
Let $b\in\{\mathrm{vis},\mathrm{geo}\}$ index the visual and geometry branches, respectively.
At layer $\ell$, each branch produces $r_b^\ell=F_b^\ell(x_t,c_b)$, where $F_b^\ell$ denotes its control embedder and ControlDiT block.
The resulting residual is linearly projected to the backbone-token shape and injected into the corresponding backbone layer~\cite{zhang2023controlnet,chen2024pixartdelta}.

\par\vspace{1pt}
\noindent\textbf{Two-stage training.}
Directly optimizing the two branches jointly leads to an imbalanced conditioning problem: the dense 3DGS rendering is highly correlated with the target video and can dominate the much sparser LiDAR signal. 
We therefore train the two branches progressively while keeping the shared VAE and pretrained video DiT frozen.
In the first stage, we optimize the geometry branch using $c_{\mathrm{lidar}}$ so that sparse 3D observations are encoded into hierarchical geometric control features.
To provide an appearance anchor, we randomly sample one frame from each video clip as $I_{\mathrm{ref}}$ and feed its image features to the LiDAR control path as an additional condition.
In the second stage, we remove $I_{\mathrm{ref}}$, freeze the trained geometry branch, and optimize the visual branch under joint $c_{\mathrm{render}}$ and $c_{\mathrm{lidar}}$ conditioning.
At this stage, the generative latent already receives stable geometric transformations and structural cues from the frozen geometry branch, allowing the visual branch to focus on supplementing appearance and suppressing reconstruction artifacts with the 3DGS rendering.
This staged optimization helps preserve the learned LiDAR geometry and reduces the tendency to copy degraded renderings as a trivial 3DGS restoration solution.
Both stages use the Flow Matching objective in Eq.~\eqref{eq:flow-matching}.

\subsection{PRG-Guided Fusion}
\label{sec:behavior}

The visual and geometry branches provide complementary controls, whose relative
contributions should vary with the local reconstruction state. We therefore use
the pixel-wise PRG label map
$L\in\{\mathrm{P},\mathrm{R},\mathrm{G}\}^{F\times H\times W}$
as a routing policy for the two control branches.
Preserve regions favor visual guidance to retain reliable 3DGS appearance;
Repair regions emphasize geometric guidance for degraded but supported content;
and Generate regions reduce reconstruction guidance to leave greater freedom
to the video generative prior.

To explicitly encode this routing policy, we define a fixed routing matrix
\begin{equation}
    \mathbf{\Lambda}
    =
    \begin{bmatrix}
        P_{\mathrm{vis}} & R_{\mathrm{vis}} & G_{\mathrm{vis}} \\
        P_{\mathrm{geo}} & R_{\mathrm{geo}} & G_{\mathrm{geo}}
    \end{bmatrix},
    \label{eq:prg-routing}
\end{equation}
where the columns correspond to Preserve, Repair, and Generate, respectively,
and the rows specify the routing strengths for the visual and geometry branches.

We first align $L$ to the video-token grid to obtain the token-level PRG map
$\widetilde{L}$. For each token $q$, its PRG state is represented as
$\mathbf{z}(q)=\operatorname{onehot}(\widetilde{L}(q))$.
The corresponding token-wise routing weights are then given by
\begin{equation}
    \begin{bmatrix}
        w_{\mathrm{vis}}(q) \\
        w_{\mathrm{geo}}(q)
    \end{bmatrix}
    =
    \mathbf{\Lambda}\mathbf{z}(q).
    \label{eq:prg-token-routing}
\end{equation}
The routing weights are shared across all control layers and modulate the branch
residuals token-wise as
$\hat r_b^\ell(q)=w_b(q)r_b^\ell(q)$ for
$b\in\{\mathrm{vis},\mathrm{geo}\}$.
The routed residuals from the two branches are subsequently fused and injected
into the corresponding frozen DiT layers.

%% file: sec/5_experiments.tex
\section{Experiments}
\label{sec:experiments}

\setcounter{dbltopnumber}{1}

\input{tables/table1_main_results}

\input{tables/table2_view_consistency}

\subsection{Experimental Setup}

\noindent\textbf{Implementation Details.}
We conduct experiments on the Waymo Open Dataset~\cite{Sun_2020_CVPR}.
Following the data curation strategy in Sec.~\ref{sec:architecture}, we process 114 scenes into 684 aligned triplets.
Each training video contains 93 frames at a resolution of $960\times640$.
We adopt Cosmos Transfer 2.5~\cite{nvidia2026cosmos} as the frozen video DiT backbone and train the control branches in two stages.
Each stage runs for 2,000 steps on 8 NVIDIA H100 GPUs with a batch size of 1 per GPU and a learning rate of $1\times10^{-5}$, for 4,000 training steps in total.

\noindent\textbf{Baselines.}
We compare PRG-Fusion against representative methods from three paradigms: reconstruction-based methods (OmniRe~\cite{chen2024omnire}), generation-based methods (FreeVS~\cite{wang2024freevs} and StreetCrafter~\cite{yan2024streetcrafter}), and restoration-based methods that enhance reconstruction with generative priors (ReconDreamer~\cite{ni2024recondreamer}, ReconDreamer++~\cite{zhao2025recondreamerplus}, Difix3D~\cite{wu2025difix3d}, and FaithFusion~\cite{wang2025faithfusion}).

\noindent\textbf{Evaluation Metrics.}
We evaluate novel trajectory synthesis from two complementary perspectives.
(1) \emph{Novel-view quality}: Following DriveDreamer4D~\cite{zhao2024drivedreamer4d}, we adopt NTA-IoU and NTL-IoU to evaluate the fidelity of generated traffic agents (e.g., vehicles and pedestrians) and lane markings, respectively, and use FID to measure overall visual quality.
(2) \emph{View consistency}: We measure distribution-level consistency between source and generated videos using FVD with R3D-18 features~\cite{unterthiner2018fvd}, and compute frame-wise CLIP similarity (CLIP-V)~\cite{radford2021clip} to evaluate semantic consistency across views.

\begin{figure}[!t]
    \centering
    \includegraphics[width=\columnwidth]{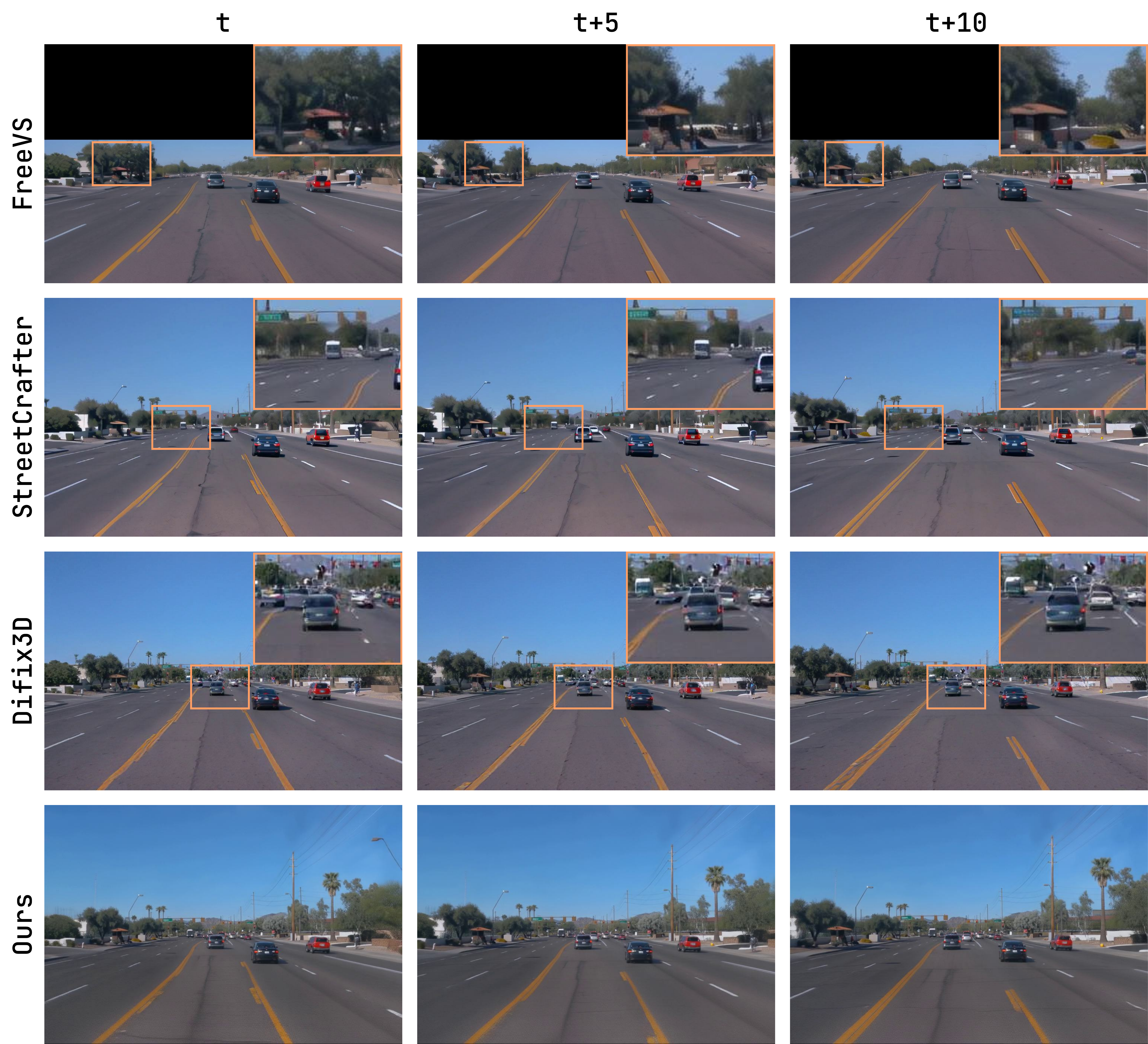}
    \caption{\textbf{Qualitative comparison of view consistency.} Note that the officially released FreeVS model is trained and evaluated at a cropped resolution that excludes the sky region to reduce computation and avoid regions with sparse LiDAR observations.}
    \label{fig:temporal-comparison}
    \vspace{-6pt}
\end{figure}

\begin{figure*}[t]
    \centering
    \includegraphics[width=\textwidth]{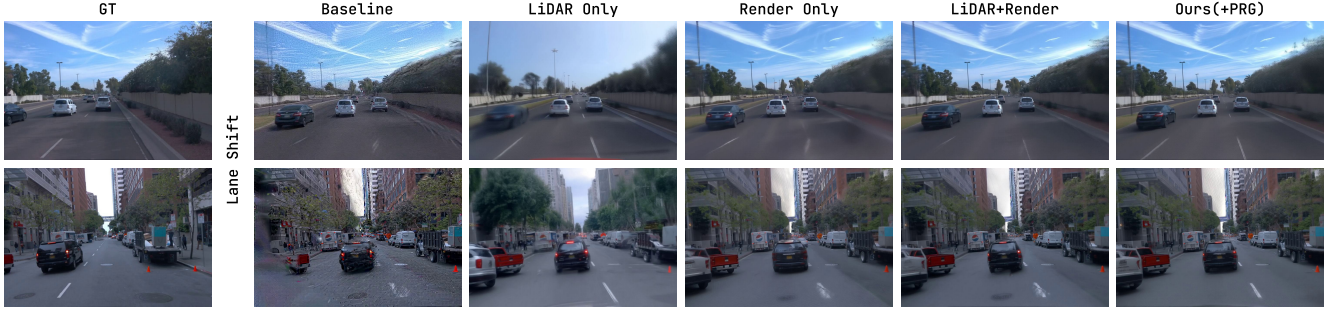}
    \caption{\textbf{Qualitative ablation of PRG-Fusion.}
    We progressively introduce LiDAR geometry, 3DGS rendering, and PRG-guided routing into the Cosmos Transfer 2.5 baseline. PRG routing enables region-adaptive coordination of geometric and visual controls, improving structural fidelity and scene appearance.}
    \label{fig:evidence-gating-ablation}
    \vspace{-8pt}
\end{figure*}

\subsection{Comparison Results}
\noindent\textbf{Quantitative comparison.}
As shown in~\cref{tab:main-results}, PRG-Fusion achieves the best NTA-IoU, NTL-IoU, and FID under the 3\,m shift.
Under the more challenging 6\,m shift, FaithFusion achieves the highest NTA-IoU, while PRG-Fusion achieves the best NTL-IoU and reduces FID by 4.36 compared with the second-best method.
The increasing FID margin at larger trajectory shifts indicates that PRG-Fusion better preserves generation quality under stronger viewpoint extrapolation.
In terms of view consistency, PRG-Fusion also achieves the best FVD and CLIP-V.
The lower FVD indicates better agreement with the recorded scene in the spatiotemporal feature distribution, while the higher CLIP-V reflects stronger preservation of scene semantics and key content across viewpoint changes.
Together, these results show that PRG-Fusion effectively suppresses scene drift while synthesizing novel-view content.

\noindent\textbf{Qualitative comparison.}
As shown in~\cref{fig:qualitative-comparison}, OmniRe suffers from prominent reconstruction artifacts and holes in newly exposed regions after trajectory shifts.
Generation-based methods can complete unseen content but tend to introduce structures inconsistent with the recorded scene, while restoration-based methods may blur lane markings or distort nearby vehicles and scene boundaries.
These failures become more pronounced under the 6\,m shift.
In contrast, PRG-Fusion better preserves vehicle appearance, lane structures, and overall scene layout under both offsets.
Furthermore, the temporal comparison in~\cref{fig:temporal-comparison} highlights its view consistency across consecutive frames: as the camera moves along the novel trajectory, PRG-Fusion consistently maintains vehicle identity and appearance, lane topology, and distant scene structure, whereas competing methods are more prone to local flickering, content drift, and abrupt structural changes across frames.

\subsection{Ablation Study}

We use Cosmos Transfer 2.5 as the baseline and progressively evaluate LiDAR projection conditioning, 3DGS rendering conditioning, their joint use, and PRG-guided fusion.
For each variant, we compute NTA-IoU and NTL-IoU on the 3\,m shift, 6\,m shift, and lane-change trajectories, and report the average across the three settings.

\suppressfloats[t]
\input{tables/table3_ablation}

\noindent\textbf{Geometry conditioning.}
Adding LiDAR conditioning substantially improves NTL-IoU while providing a more moderate gain in NTA-IoU.
As shown in~\cref{fig:evidence-gating-ablation}, LiDAR guidance better preserves road geometry and lane structure, while its sparse observations provide limited appearance information.
This confirms its primary role as a reliable geometric constraint.

\noindent\textbf{Visual conditioning.}
Conditioning on 3DGS renderings instead yields a pronounced improvement in NTA-IoU and also benefits NTL-IoU.
The corresponding results better preserve scene-specific appearance and vehicle details, although geometric boundaries such as lane markings and object contours remain less accurate.

\noindent\textbf{Joint conditioning.}
Combining LiDAR geometry with 3DGS appearance further improves NTA-IoU while retaining a clear NTL-IoU gain over the baseline.
The joint model preserves richer appearance than geometry conditioning alone and stronger scene structure than visual conditioning alone.
However, uniform fusion applies the same control strategy across spatially varying reconstruction states, which can introduce structural inconsistencies where the two cues conflict.

\noindent\textbf{PRG-guided fusion.}
PRG-guided fusion achieves the best NTA-IoU and NTL-IoU among all variants.
Its improvement over uniform joint conditioning shows that the gains arise not only from combining visual and geometric cues, but also from adapting their contributions to local reconstruction states.
The qualitative results in~\cref{fig:evidence-gating-ablation} show a consistent trend, with clearer lane structures and object boundaries while preserving vehicle and scene appearance.
This further supports the effectiveness of region-aware control.

%% file: tables/table1_main_results.tex
\begin{table*}[t]
    \centering
    \small
    \setlength{\tabcolsep}{6pt}
    \renewcommand{\arraystretch}{1.12}
    \begin{tabular*}{\textwidth}{@{\extracolsep{\fill}}lcccccc@{}}
        \toprule
        & \multicolumn{3}{c}{Shift 3 m} & \multicolumn{3}{c}{Shift 6 m} \\
        \cmidrule(lr){2-4} \cmidrule(lr){5-7}
        Method & NTA-IoU $\uparrow$ & NTL-IoU $\uparrow$ & FID $\downarrow$ & NTA-IoU $\uparrow$ & NTL-IoU $\uparrow$ & FID $\downarrow$ \\
        \midrule
        OmniRe & 0.424 & 51.73 & 188.42 & 0.423 & 49.08 & 191.00 \\
        \addlinespace[2pt]
        FreeVS & 0.505 & 56.84 & 104.23 & 0.465 & 55.37 & 121.44 \\
        StreetCrafter & 0.540 & 55.54 & 92.29 & 0.483 & 54.87 & 116.71 \\
        \addlinespace[2pt]
        ReconDreamer & 0.539 & 54.58 & 93.56 & 0.467 & 52.58 & 149.19 \\
        ReconDreamer++ & 0.572 & 57.06 & 72.02 & 0.489 & \underline{56.57} & 111.92 \\
        Difix3D & 0.578 & 56.94 & 84.12 & \underline{0.504} & 53.77 & 120.24 \\
        FaithFusion & \underline{0.581} & \underline{57.67} & \underline{71.51} & \textbf{0.517} & 55.78 & \underline{107.47} \\
        \midrule
        \textbf{Ours} & \textbf{0.584} & \textbf{58.72} & \textbf{69.43} & 0.502 & \textbf{58.12} & \textbf{103.11} \\
        \bottomrule
    \end{tabular*}
    \caption{\textbf{Quantitative comparison under 3 m and 6 m lateral trajectory shifts.} The best and second-best results are shown in bold and underlined, respectively.}
    \label{tab:main-results}
    \vspace{-9pt}
\end{table*}

%% file: tables/table2_view_consistency.tex
\begin{table}[t]
    \centering
    \small
    \setlength{\tabcolsep}{5.5pt}
    \renewcommand{\arraystretch}{1.12}
    \begin{tabular}{@{}lcc@{}}
        \toprule
        Method & FVD $\downarrow$ & CLIP-V $\uparrow$ \\
        \midrule
        StreetCrafter & 41.78 & \underline{98.11} \\
        FreeVS & 96.62 & 97.24 \\
        Difix3D & \underline{41.30} & 97.68 \\
        \textbf{Ours} & \textbf{40.66} & \textbf{98.36} \\
        \bottomrule
    \end{tabular}
    \caption{\textbf{Quantitative comparison of view consistency.}}
    \label{tab:view-consistency}
    \vspace{-9pt}
\end{table}

%% file: tables/table3_ablation.tex
\begin{table}[t]
    \centering
    \small
    \setlength{\tabcolsep}{5pt}
    \renewcommand{\arraystretch}{1.12}
    \begin{tabular}{@{}lcc@{}}
        \toprule
        Variant & NTA-IoU $\uparrow$ & NTL-IoU $\uparrow$ \\
        \midrule
        Baseline & 0.449 & 53.06 \\
        LiDAR only & 0.476 & \underline{57.85} \\
        Render only & 0.513 & 55.68 \\
        LiDAR + Render & \underline{0.539} & 56.39 \\
        \textbf{Ours (+ PRG)} & \textbf{0.543} & \textbf{57.94} \\
        \bottomrule
    \end{tabular}
    \caption{\textbf{Ablation of input evidence and PRG gating.}}
    \label{tab:evidence-gating-ablation}
    \vspace{-8pt}
\end{table}

%% file: sec/6_conclusion.tex
\section{Conclusion}
\label{sec:conclusion}

In this work, we presented PRG-Fusion, a region-aware video generation framework for novel trajectory synthesis in autonomous driving. By converting spatially varying reconstruction evidence into pixel-wise Preserve, Repair, and Generate (PRG) labels, PRG-Fusion adaptively coordinates dense 3DGS appearance, sparse LiDAR geometry, and video generative priors according to the local reconstruction state. A dual-branch control architecture with two-stage progressive training further promotes complementary visual and geometric guidance while mitigating over-reliance on degraded 3DGS renderings. Extensive experiments demonstrate that PRG-Fusion achieves improved visual quality, geometric fidelity, and view consistency.

\noindent\textbf{Limitations and Future Work.}
The current framework still relies on scene-specific 3DGS reconstruction and LiDAR geometry, introducing additional preprocessing overhead. Future work will explore feed-forward reconstruction models and predicted depth as alternatives to improve scalability and reduce sensor dependence.

%% file: sec/7_appendix.tex

\section{Additional PRG Extraction Details}
\label{app:prg}

This section provides additional implementation details for the reconstruction
evidence and PRG assignment introduced in Sec.~\ref{sec:features}.
Given a reconstructed 3DGS scene $\mathcal G=\{G_i\}_{i=1}^{N}$, the training
views $\mathcal V^{\mathrm{train}}$, and a target camera trajectory
$\mathcal V^{\mathrm{tar}}$, we construct a pixel-wise PRG label sequence
$L_f\in\{\mathrm P,\mathrm R,\mathrm G\}^{H\times W}$.

\subsection{Gaussian-Level Evidence}

\noindent\textbf{Observation support.}
For each Gaussian $G_i$, we estimate its reconstruction support from visibility
frequency and viewing-direction coverage.
Let $n_i$ denote its number of visible training views.
We first compute a robust visibility score
\begin{equation}
    v_i=
    \operatorname{clip}\!\left(
    \frac{\log(1+n_i)}{q_v},0,1
    \right),
    \label{eq:supp-visibility-score}
\end{equation}
where $q_v$ is the 95th percentile of $\log(1+n_j)$ over visible Gaussians.
The logarithm suppresses long-tailed observation counts.

We further cluster the observed viewing directions into $K_d=5$ prototypes.
If $G_i$ occupies $k_i$ prototypes, its directional coverage is
$d_i=k_i/K_d$. We use directional diversity to modulate the visibility
evidence:
\begin{equation}
    s_i=
    v_i\big[(1-\lambda_d)+\lambda_d d_i\big],
    \label{eq:supp-observation-support}
\end{equation}
where $\lambda_d=0.35$.
Thus, visibility frequency provides the primary evidence, while directional
coverage provides a bounded modulation.

\noindent\textbf{Trajectory-dependent unreliability.}
Following FaithFusion~\cite{wang2025faithfusion}, we use rendering-gradient
sensitivity to measure how reliably each Gaussian is supported when observed
from the target trajectory.
For Gaussian parameters $\omega_i$, the accumulated training response is
\begin{equation}
    H_i^{\mathrm{train}}
    =
    \sum_k
    \left\lVert
    \nabla_{\omega_i}
    \mathcal R(V_k^{\mathrm{train}})
    \right\rVert_1.
    \label{eq:supp-training-response}
\end{equation}
Sky gradients are masked, and each pixel contribution is normalized by the
number of Gaussians intersecting the corresponding ray.

For target frame $f$, we compute the corresponding target response
$H_{i,f}^{\mathrm{tar}}$ and obtain
\begin{equation}
    e_{i,f}
    =
    \frac{H_{i,f}^{\mathrm{tar}}}
    {H_i^{\mathrm{train}}+\lambda},
    \qquad \lambda=10^{-6}.
    \label{eq:supp-raw-eig}
\end{equation}
A larger response indicates that the target rendering depends strongly on a
Gaussian that is weakly constrained by the training observations.

Because response magnitudes vary across target frames, we robustly normalize
them using the 5th and 95th percentiles:
\begin{equation}
    \widetilde e_{i,f}
    =
    \operatorname{clip}\!\left(
    \frac{e_{i,f}-a_f}{b_f-a_f},0,1
    \right).
    \label{eq:supp-eig-normalization}
\end{equation}

For a frame set $\mathcal A$, we summarize the normalized responses by their
mean $\mu_i^{\mathcal A}$, fourth-order generalized mean
$r_i^{\mathcal A}$, and the frequency $q_i^{\mathcal A}$ of responses above
$0.5$. These statistics capture persistent, strong, and recurrent
unreliability, respectively. We combine them as
\begin{equation}
    u_i^{\mathcal A}
    =
    0.25\mu_i^{\mathcal A}
    +0.60r_i^{\mathcal A}
    +0.15q_i^{\mathcal A}.
    \label{eq:supp-eig-aggregation}
\end{equation}

We evaluate this statistic both globally over the full trajectory and locally
using 32-frame windows with a stride of 16. The final Gaussian-level
unreliability is
\begin{equation}
    \bar u_i(f)
    =
    0.30u_i^{\mathrm{scene}}
    +0.70u_i^{\mathrm{local}}(f),
    \label{eq:supp-temporal-unreliability}
\end{equation}
where the global term stabilizes the estimate and the local term captures
changes along the target trajectory.

\subsection{Pixel-Level Evidence and PRG Assignment}

\noindent\textbf{Geometric support.}
Observation statistics alone cannot determine whether a surface exposed in
the target view is geometrically supported by the recorded observations.
We therefore back-project each target pixel using its rendered depth and
reproject the resulting 3D point into the training views.
For a valid projection into training view $k$, we measure the relative depth
discrepancy as
\begin{equation}
    \delta_{f,p}^{k}
    =
    \frac{|D_k-z_{f,p}^{k}|}
    {\max(|D_k|,|z_{f,p}^{k}|,\epsilon_d)},
    \label{eq:supp-relative-depth-error}
\end{equation}
where $z_{f,p}^{k}$ is the projected point depth and $D_k$ is the rendered
training-view depth at the same location.

We convert the discrepancy into a soft consistency score
\begin{equation}
    g_{f,p}^{k}
    =
    \exp\!\left(-\delta_{f,p}^{k}/\tau_d\right),
    \label{eq:supp-depth-consistency}
\end{equation}
and retain the strongest valid support:
\begin{equation}
    G_f(p)=\max_k g_{f,p}^{k}.
    \label{eq:supp-geometric-support}
\end{equation}
Invalid projections, non-positive depths, and invalid-opacity pixels receive
zero support. We use $\tau_d=0.10$ and $\epsilon_d=10^{-3}$, and evaluate this
cross-view consistency at four-times downsampled spatial resolution.

\noindent\textbf{Evidence projection.}
We project Gaussian-level unreliability and observation support to the target
view using scalar alpha compositing:
\begin{equation}
    U_f(p)=\Pi_f(\{\bar u_i(f)\})(p),
    \qquad
    \widehat S_f(p)=\Pi_f(\{s_i\})(p),
    \label{eq:supp-evidence-projection}
\end{equation}
where $\Pi_f$ denotes the standard 3DGS compositing operator with the Gaussian
color replaced by the corresponding scalar evidence.
The final reconstruction support additionally requires geometric consistency:
\begin{equation}
    S_f(p)=\widehat S_f(p)G_f(p).
    \label{eq:supp-trusted-support}
\end{equation}

\noindent\textbf{PRG assignment.}
PRG assignment follows a hierarchical decision rule.
Unreliability first determines whether the current reconstruction should be
preserved, while reconstruction support further separates Repair from Generate.
Let $X_f(p)$ denote the exclusion mask for sky and the ego vehicle.
Using $\tau_u=\tau_s=0.25$, we assign
\begin{equation}
    L_f(p)=
    \begin{cases}
        \mathrm P,
        & X_f(p)=1\ \text{or}\ U_f(p)<\tau_u,\\
        \mathrm R,
        & U_f(p)\geq\tau_u\ \text{and}\ S_f(p)\geq\tau_s,\\
        \mathrm G,
        & U_f(p)\geq\tau_u\ \text{and}\ S_f(p)<\tau_s.
    \end{cases}
    \label{eq:supp-prg-assignment}
\end{equation}
Thus, reliable pixels are preserved, unreliable but sufficiently supported
pixels are assigned to Repair, and weakly supported pixels are assigned to
Generate.


\section{Implementation Details}
\label{app:implementation}

\subsection{Training Data and Condition Construction}

For each training scene, we first reconstruct a 3DGS representation from the
recorded driving log. Along each target camera trajectory, we render the
reconstructed scene to obtain the visual condition
$c_{\mathrm{render}}$.
We aggregate and colorize LiDAR observations and project them onto the same
target camera sequence to obtain the geometry condition
$c_{\mathrm{lidar}}$.
The synchronized RGB video provides the training target, resulting in aligned
$(c_{\mathrm{render}},c_{\mathrm{lidar}},V)$ triplets.

We process 114 Waymo scenes into 684 aligned training triplets.
The two control branches are trained following the two-stage progressive
strategy described in Sec.~\ref{sec:architecture}.

\subsection{Training and Inference Settings}

We use Cosmos Transfer 2.5~\cite{nvidia2026cosmos} as the frozen video DiT
backbone. Each video clip contains 93 frames at 10 fps with a spatial
resolution of $960\times640$.
The two training stages run for 2,000 steps each, for 4,000 steps in total.
The LiDAR condition aggregates five sweeps from five cameras.

\Cref{tab:supp-hyperparameters} summarizes the remaining settings.

\begin{table}[h]
    \centering
    \small
    \setlength{\tabcolsep}{5pt}
    \begin{tabular}{@{}lc@{}}
        \toprule
        Setting & Value \\
        \midrule
        Video input & $93$ frames, $10$ fps, $960\times640$ \\
        LiDAR aggregation & 5 sweeps, 5 cameras \\
        Masked LiDAR rows & Top 256 \\
        Iterations per stage & 2,000 \\
        Hardware / batch size & 8 H100s / 1 per GPU \\
        Optimizer & AdamW \\
        Learning rate & $1\times10^{-5}$ \\
        Warm-up & 100 iterations \\
        Weight decay & $10^{-3}$ \\
        Gradient clipping & 0.1 \\
        Numerical precision & BF16 network, FP32 flow \\
        \bottomrule
    \end{tabular}
    \caption{\textbf{Training and inference settings.}}
    \label{tab:supp-hyperparameters}
\end{table}

\subsection{PRG Routing Parameters}

The PRG thresholds and routing parameters are fixed globally and shared across
all evaluation scenes and control layers.
We use
\begin{equation}
    (\tau_u,\tau_s)=(0.25,0.25),
\end{equation}
and the routing matrix
\begin{equation}
    \mathbf{\Lambda}
    =
    \begin{bmatrix}
        1.00 & 0.00 & 0.50 \\
        0.50 & 0.25 & 0.00
    \end{bmatrix},
    \label{eq:supp-routing-matrix}
\end{equation}
where the first and second rows correspond to the visual and geometry branches,
and the columns correspond to Preserve, Repair, and Generate, respectively.


\section{Evaluation Protocol}
\label{app:evaluation-protocol}

\subsection{Evaluation Scenes}
\label{app:scene-selection}

We evaluate on eight scenes from the validation split of the Waymo Open
Dataset~\cite{Sun_2020_CVPR} containing substantial dynamic-agent motion.
The sequence identifiers and evaluation frame ranges are reported in
\cref{tab:evaluation-scenes} to ensure reproducibility.

\begin{table}[t]
    \centering
    \caption{\textbf{Evaluation scenes.} Unique identifiers of the selected Waymo validation segments and their inclusive frame ranges.}
    \label{tab:evaluation-scenes}
    \scriptsize
    \setlength{\tabcolsep}{3pt}
    \renewcommand{\arraystretch}{1.06}
    \begin{tabular}{@{}lc@{}}
        \toprule
        Waymo segment ID & Frames \\
        \midrule
        \texttt{\detokenize{10359308928573410754_720_000_740_000}} & 120--159 \\
        \texttt{\detokenize{12820461091157089924_5202_916_5222_916}} & 0--39 \\
        \texttt{\detokenize{15021599536622641101_556_150_576_150}} & 0--39 \\
        \texttt{\detokenize{16767575238225610271_5185_000_5205_000}} & 0--39 \\
        \texttt{\detokenize{17152649515605309595_3440_000_3460_000}} & 60--99 \\
        \texttt{\detokenize{17860546506509760757_6040_000_6060_000}} & 90--129 \\
        \texttt{\detokenize{2506799708748258165_6455_000_6475_000}} & 80--119 \\
        \texttt{\detokenize{3015436519694987712_1300_000_1320_000}} & 40--79 \\
        \bottomrule
    \end{tabular}
\end{table}

\begin{figure*}[t]
    \centering
    \includegraphics[width=\textwidth]{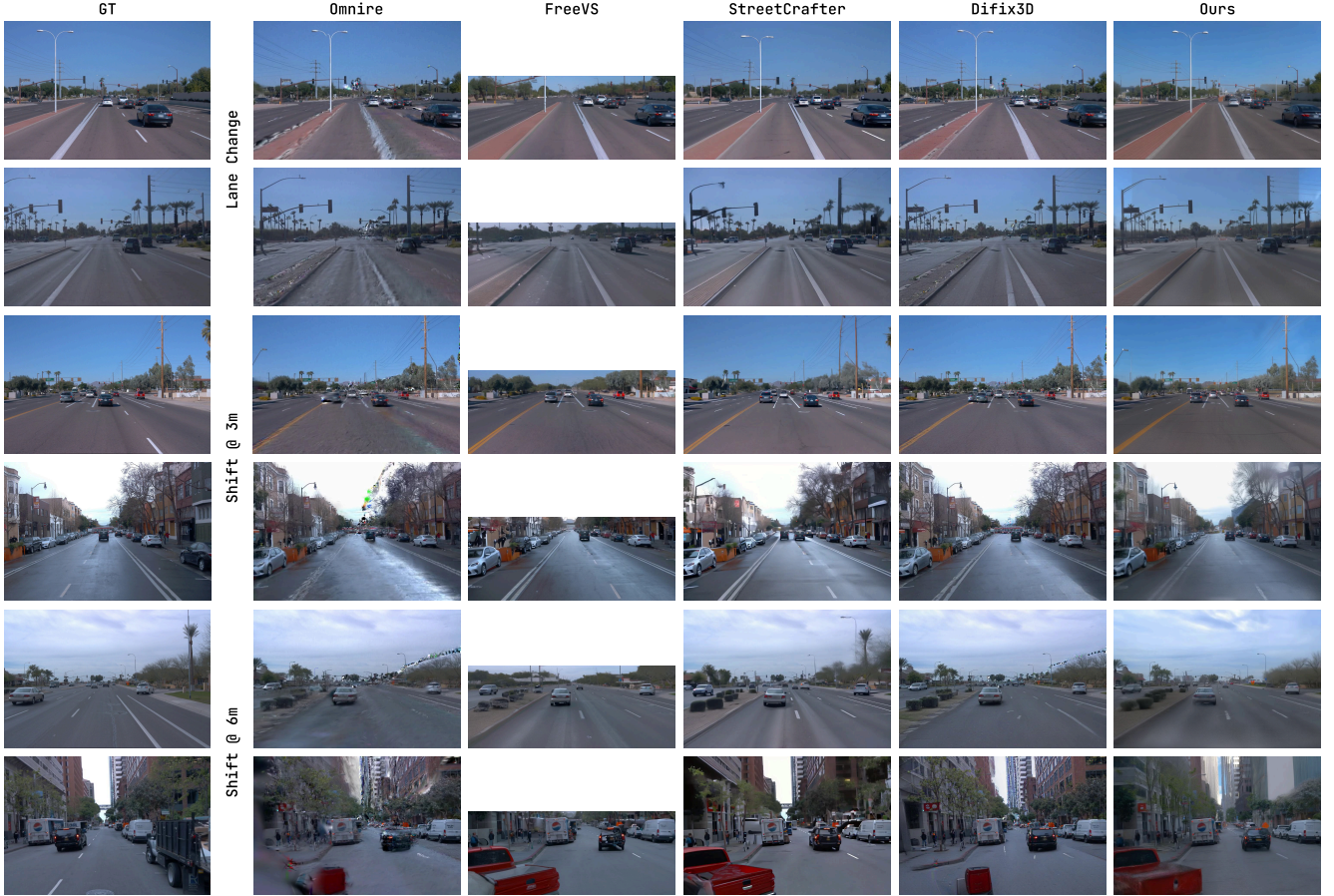}
    \caption{\textbf{Additional qualitative comparisons across novel trajectories.}
    Rows show representative examples from lane-change, 3\,m lateral-shift, and 6\,m lateral-shift settings.
    The column marked GT is the recorded-view reference from the original camera trajectory, rather than pixel-aligned RGB ground truth at the target novel viewpoint.
    FreeVS is shown at its officially released cropped resolution, which excludes the sky region.}
    \label{fig:supp-qualitative}
    \vspace{-6pt}
\end{figure*}

\subsection{Novel-View Quality Metrics}
\label{app:novel-view-metrics}

Following DriveDreamer4D~\cite{zhao2024drivedreamer4d}, we evaluate novel-view quality from both geometric and perceptual perspectives.
Because real RGB observations are unavailable at the shifted viewpoints, the source-trajectory images are not used as pixel-level ground truth for the two IoU metrics.

\noindent\textbf{Novel Trajectory Agent IoU (NTA-IoU).}
NTA-IoU measures the geometric fidelity of foreground traffic agents.
We detect 2D bounding boxes in each synthesized frame and compare them with the ground-truth Waymo 3D bounding boxes projected into the same target viewpoint.
Higher NTA-IoU indicates better preservation of agent positions and spatial extents.

\noindent\textbf{Novel Trajectory Lane IoU (NTL-IoU).}
NTL-IoU measures the geometric fidelity of lane structures.
We detect lane lines in each synthesized frame and compare them with the Waymo HDMap lane geometry projected into the target viewpoint.
Higher NTL-IoU indicates better preservation of static road geometry.

\noindent\textbf{Fr\'echet Inception Distance (FID).}
FID evaluates frame-level visual quality by comparing the Inception feature distributions of synthesized frames and real frames from the evaluation scenes.
This distribution-level metric does not require pixel-aligned RGB ground truth at the shifted viewpoints, and lower FID indicates higher visual fidelity.

\subsection{View Consistency Metrics}
\label{app:view-consistency-metrics}

We further evaluate whether the synthesized videos remain consistent with the recorded scene across viewpoint changes at both video-distribution and semantic levels.

\noindent\textbf{FVD-R3D.}
FVD-R3D measures video-level distribution consistency using R3D-18 features~\cite{unterthiner2018fvd}.
We extract clip-level features from source-trajectory and synthesized videos and compute the Fr\'echet distance between the two distributions.
Lower FVD-R3D indicates better preservation of scene content and temporal dynamics.

\noindent\textbf{Video CLIP Similarity (CLIP-V).}
CLIP-V measures cross-view semantic consistency using CLIP image features~\cite{radford2021clip}.
We compute the cosine similarity between temporally corresponding source and synthesized frames and average the scores over all frames and scenes.
Higher CLIP-V indicates stronger preservation of scene semantics under viewpoint changes.


\section{Additional Qualitative Results}
\label{app:additional-qualitative}

Additional comparisons across all three novel-trajectory settings are shown in Fig.~\ref{fig:supp-qualitative}.